\pdfoutput=1 

\documentclass[11pt]{article}

\usepackage{ACL2023}

\usepackage{times}
\usepackage{latexsym}

\usepackage[T1]{fontenc}

\usepackage[utf8]{inputenc}

\usepackage{microtype}

\usepackage{inconsolata}

\usepackage{graphicx}
\usepackage{pifont}
\usepackage{multirow}
\usepackage{array}
\usepackage{dblfloatfix}
\usepackage{titlesec}
\titlespacing*{\section}
{0pt}{5.2ex plus 1ex minus .2ex}{1.4ex plus .2ex}
\newcommand\blfootnote[1]{%
  \begingroup
  \renewcommand\thefootnote{}\footnote{#1}%
  \addtocounter{footnote}{-1}%
  \endgroup
}

\newcommand{\scifive}{\textsc{SciFive}}
\newcommand{\tfive}{\textsc{T5}}
\newcommand{\flantfive}{\textsc{FLAN-T5}}
\newcommand{\clintfive}{\textsc{Clin-T5}}
\newcommand{\clintfivesci}{\textsc{Clin-T5-Sci}}

\newcommand{\clintfiveb}{\textsc{Clin-T5-Base}}

\newcommand{\clintfivel}{\textsc{Clin-T5-Large}}

\definecolor{r012_Q1}{RGB}{72,0,183}
\definecolor{r012_Q2}{RGB}{62,0,193}
\definecolor{r012_Q3}{RGB}{16,0,239}
\definecolor{r0_Q1}{RGB}{102,0,153}
\definecolor{r0_Q2}{RGB}{62,0,193}
\definecolor{r0_Q3}{RGB}{51,0,204}
\definecolor{r1_Q1}{RGB}{62,0,193}
\definecolor{r1_Q2}{RGB}{62,0,193}
\definecolor{r1_Q3}{RGB}{0,0,255}
\definecolor{r2_Q1}{RGB}{51,0,204}
\definecolor{r2_Q2}{RGB}{57,0,198}
\definecolor{r2_Q3}{RGB}{0,0,255}

\title{RadAdapt: Radiology Report Summarization via Lightweight\\Domain Adaptation of Large Language Models}

\author{Dave Van Veen*, Cara Van Uden*, Maayane Attias, {\bf Anuj Pareek},  \\ {\bf Christian Bluethgen}, 
         {\bf Malgorzata Polacin}, {\bf Wah Chiu}, \\ 
        {\bf Jean-Benoit Delbrouck}, {\bf Juan Manuel Zambrano Chaves}, \\
        {\bf Curtis P. Langlotz}, {\bf Akshay S. Chaudhari}, {\bf John Pauly}  \\
        Stanford University \\ \texttt{\{vanveen, cvanuden\}@stanford.edu}}

\begin{document}
\maketitle

\def\thefootnote{*}\footnotetext{Equal contribution}\def\thefootnote{\arabic{footnote}}

\begin{abstract}
    We systematically investigate lightweight strategies to adapt large language models (LLMs) for the task of radiology report summarization (RRS).
    Specifically, we focus on domain adaptation via pretraining (on natural language, biomedical text, or clinical text) and via discrete prompting or parameter-efficient fine-tuning.
    Our results consistently achieve best performance by maximally adapting to the task via pretraining on clinical text and fine-tuning on RRS examples. Importantly, this method fine-tunes a mere 0.32\% of parameters throughout the model, in contrast to end-to-end fine-tuning (100\% of parameters).
    Additionally, we study the effect of in-context examples and out-of-distribution (OOD) training before concluding with a radiologist reader study and qualitative analysis. Our findings highlight the importance of domain adaptation in RRS and provide valuable insights toward developing effective natural language processing solutions for clinical tasks.
\end{abstract}
\section{Introduction}

\blfootnote{Code: \url{https://github.com/davevanveen/radadapt}}
Radiology reports are comprehensive documents that capture and interpret the results of a radiological imaging examination. Reports are often structured into three main sections: (1) a \textit{background} section that provides general information about the exam and the patient, e.g.~medical history (2) a \textit{findings} section that presents detailed exam analysis and results, and (3) an \textit{impression} section that concisely summarizes the most salient findings. In a typical workflow, a radiologist first dictates the detailed findings and then distills them into a concise impression. This impression is the most significant part of a radiology report, as it contains crucial information for clinical decision-making~\cite{kahn2009toward}. However, performing radiology report summarization (RRS) manually can be labor-intensive and prone to errors~\cite{gershanik2011critical}, motivating the importance of automating this task.

Large language models (LLMs) have demonstrated remarkable capabilities in natural language understanding and generation, serving as foundation models that can be adapted to various domains and tasks. However, their sheer size, sometimes exceeding 100B parameters, makes training for domain-specific tasks prohibitively expensive in terms of computation and training data. We address this challenge by exploring lightweight strategies for domain adaptation in the context of RRS, culminating in the following contributions:

 \begin{itemize}
    \item We systematically evaluate a variety of LLMs and lightweight adaptation methods, achieving the best performance by pretraining on clinical text and performing parameter-efficient fine-tuning with LoRA~\cite{hu2021lora}. Generated impressions showcase the effectiveness of lightweight adaptation strategies for RRS.
    \item We investigate the impact of few-shot prompting by conducting ablation studies on the number of in-context examples provided to each model. Our findings reveal that increased context leads to improved performance across almost all cases, shedding light on the value of prompt engineering when adapting LLMs for RRS.
    \item We evaluate ``out-of-distribution'' (OOD) model performance and specifically examine the model's ability to generalize to different imaging modalities and anatomies. Our results indicate that anatomy plays a more crucial role than modality, and best performance is achieved when training on a larger dataset which encompasses all modalities and anatomies.
    \item We conduct a reader study with radiologists who provide qualitative insights and quantitative scores on the model's correctness, coherence, and ability to capture critical information. While our results are promising, we emphasize the need for further improvements and evaluation before clinical deployment.
\end{itemize}

Overall, our research presents a comprehensive investigation of lightweight strategies for domain adaptation in the context of RRS, offering insights into the effectiveness of different approaches and highlighting the potential of LLMs for this task. Our findings contribute toward the advancement of applied natural language processing (NLP) to radiology with implications for improving radiologists' workflow and patient care.
\section{Related Work}

In recent years, transformer-based~\cite{vaswani2017attention} language models have become ubiquitous in NLP due to their state-of-the-art performance across many tasks including language generation, question answering, and machine translation. Transformer models BERT~\cite{devlin2018bert} and GPT-2~\cite{radford2019language} established a new paradigm of first training on large amounts of general data and then fine-tuning on domain-specific data, as opposed to direct training on domain-specific data. This has led to training transformers with more parameters on increasingly more data, resulting in LLMs such as GPT-3~\cite{brown2020language}, PaLM~\cite{chowdhery2022palm}, and the ``text-to-text transfer transformer,'' or T5~\cite{raffel2020t5}. 

However, end-to-end fine-tuning LLMs like GPT-3 (175B parameters) requires substantial computational resources, creating a high barrier to entry.
As a result, there has been a growing interest in lightweight methods for domain adaptation. One such method is prompting, in which one provides initial text input to the LLM so it has context for the given task. Performance depends heavily on the provided prompt~\cite{brown2020language}, motivating principled prompting methods. Various works have pursued this in the form of natural language instructions~\cite{liu2023prompt,wei2022chain} or supplying ``in-context'' examples of desired output \cite{lampinen2022can}. An alternative approach consists of parameter-efficient fine-tuning, where one freezes existing model weights and inserts a small number of tunable parameters~\cite{rebuffi2017learning, houlsby2019parameter, lin2020exploring}. We focus on the two highest regarded parameter-efficient fine-tuning methods: prefix tuning~\cite{li2021prefix,lester2021power} and LoRA~\cite{hu2021lora}, discussed further in Section~\ref{sec:methods}.

In addition to methods for prompting and fine-tuning, previous work has demonstrated adaptation to the medical domain via pretraining on biomedical or clinical text. Consider SciFive~\cite{phan2021scifive}, Clinical-T5~\cite{lehman2023clint5}, and Med-PaLM~\cite{singhal2022large}, which leveraged LLMs for text generation on medical tasks. Both SciFive (biomedical) and Clinical-T5 (clinical) achieved state-of-the art results for their respective domains in tasks such as named entity recognition, natural language inference, and question-answering. Additionally, Med-PaLM achieved success aligning text generation models to clinical tasks via methods such as chain-of-thought prompting, prompt tuning~\cite{lester2021power}, and instruction tuning~\cite{chung2022flant5}.

Within the clinical domain, we focus specifically on the task of RRS. Previous work has approached this task with a focus on consistency and factual correctness~\cite{zhang2020,dai2021bdkg,miura2021,delbrouck2022radgraph}. To the best of our knowledge, this work is the first to leverage lightweight domain adaptation strategies on LLMs for the RRS task. Additionally, most prior work on RRS uses only chest x-rays~\cite{dai2021bdkg,abacha2021overview} from large datasets like MIMIC-CXR~\cite{johnson2019mimiccxr} and CheXpert~\cite{irvin2019chexpert}. In contrast, our work employs the MIMIC-III dataset~\cite{johnson2016mimic}, which contains longer radiology reports from a diverse set of two imaging modalities (MR, CT) and seven anatomies (head, chest, etc.), presenting a more difficult summarization task.
\section{Methods}\label{sec:methods}
\begin{figure*}[ht!]
    \centering
    \includegraphics[width = 1\textwidth]{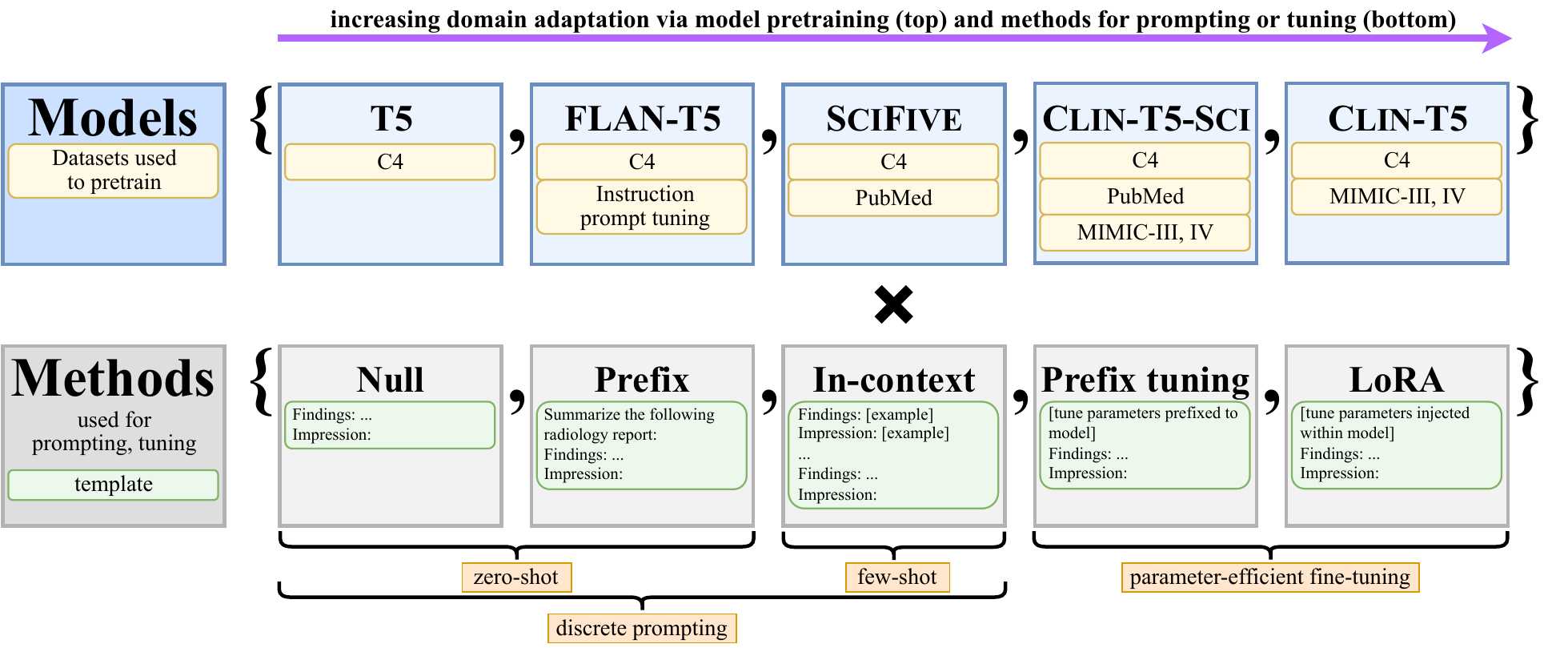}
    \caption{Diagram of experiments. We evaluate every combination of pretrained LLM (top) and lightweight adaptation method (bottom). Moving from left to right, the models and methods become increasingly adapted to the downstream clinical task of RRS.}
    \label{fig:experiments}
\end{figure*}

As depicted in Figure~\ref{fig:experiments}, we investigate adapting LLMs to the task of RRS along two axes: (1) models pretrained on various combinations of natural language, biomedical, and clinical text data, as described in Section~\ref{label:models}, and (2) various methods for discrete prompting and parameter-efficient fine-tuning, as described in Section~\ref{label:adaptation}.

\subsection{Pretrained Models}
\label{label:models}

To mitigate variance introduced by different model architectures, we focus this study on \tfive, i.e.~``text-to-text transfer transformer,'' a highly regarded encoder-decoder architecture available for public use~\cite{raffel2020t5}. \tfive's text-to-text framework enables the model to be used on any NLP task, and its pretraining over the C4 (Colossal Clean Crawled Corpus) dataset~\cite{raffel2020t5} enables excellent performance in transfer learning. We include results for two architecture sizes: base~(223M parameters), and large~(738M parameters). From hereon we refer to \tfive~as the original model pretrained on C4 alone.
The remaining four models are simply a version of \tfive~that was subsequently end-to-end fine-tuned on datasets of various relevance to the RRS task:
\begin{itemize}
    \item \flantfive~\cite{chung2022scaling} tuned via instruction prompt tuning.
    \item \scifive~\cite{phan2021scifive} tuned on the biomedical text dataset of PubMed, i.e.~Pubmed Abstract~\cite{pubmed} and PubMed Central~\cite{pmc}.
    \item \clintfivesci~or Clinical-T5-Sci~\cite{lehman2023clint5} tuned on PubMed and two clinical text datasets (MIMIC-III~\cite{johnson2016mimic} and MIMIC-IV~\cite{johnson2020mimic}).
    \item \clintfive~or Clinical-T5~\cite{lehman2023clint5} tuned on MIMIC-III and IV alone.
\end{itemize}
Please see the top row in Figure~\ref{fig:experiments} for a visual illustration of these five models. We acknowledge the difficulty of ranking pretraining datasets' relevance for a particular downstream task. In the case of RRS, it may seem reasonable to assert that clinical text is more relevant than biomedical text, and that biomedical text is more relevant than general natural language text. However, it becomes more difficult to compare \flantfive's instruction tuning, which drastically improves performance on prompting benchmarks but has not yet been explored for medical tasks \cite{longpre2023flan}. 
We also note that \clintfivesci~and \clintfive~were trained on MIMIC-III, the primary dataset used for evaluation. Please refer to Section~\ref{sec:results-pitfalls} for discussion and experiments addressing this complication.

\setlength{\tabcolsep}{6pt} 

\begin{table*}[ht]
\caption{We employ parameter-efficient fine-tuning methods for domain adaptation that modify <0.4\% of model parameters while keeping other parameters frozen.
}
\centering
\begin{tabular}{l l | c c | c c c}
\hline
 & & \multicolumn{2}{|c|}{\textbf{Tunable parameters}} & \multicolumn{2}{c}{\textbf{Training time (hr)}}  \\
\textbf{Model size} & \textbf{Method} & \textbf{\#} & \textbf{\% of total} & \textbf{per epoch} & \textbf{total} & \textbf{\# epochs}   \\
\hline
\multirow{2}{*}{Base (223M)} & prefix tuning & 0.37M & 0.17\% & 0.98 & 9.83 & 10  \\
& LoRA & 0.88M & 0.39\% & 1.32 & 6.60 & 5 \\
\multirow{2}{*}{Large (738M)} & prefix tuning & 0.98M & 0.13\% & 2.93 & 29.3 & 10  \\
& LoRA & 2.4M & 0.32\% & 3.85 & 19.3 & 5 \\
\end{tabular}

\label{tab:num-params}
\end{table*}

\subsection{Lightweight task adaptation methods}
\label{label:adaptation}
For each pretrained model discussed in Section~\ref{label:models}, we evaluate five lightweight domain adaptation methods for prompting and tuning (Fig.~\ref{fig:experiments}, bottom row). Prompting provides adaptation by simply supplying particular tokens to the frozen pretrained model. In contrast, parameter-efficient fine-tuning provides adaptation by adding a small number of parameters to the model and optimizing them for the task, while the original model parameters remain frozen. Compared to updating all model parameters via end-to-end fine-tuning, parameter-efficient methods require much less computation and training data. We describe these five methods below in order of increasing adaptation to the downstream RRS task:

\begin{enumerate}
    \item \textit{Null prompting}~\cite{zhao2021discrete} is a simple discrete (sequence of real natural-language tokens), zero-shot prompt. We supply the radiology report findings section and the basic prompt, ``impression:''.
    
    \item \textit{Prefixed prompting}~\cite{zhao2021discrete} is a discrete, zero-shot prompt with a brief instruction prepended to the original null prompt above. For our instruction we use ``summarize the following radiology report:''. This provides the model some context for the RRS task. 
    We note that a slight modification to the prepended instruction may significantly change the generated output for an individual sample. However, that same modification does not meaningfully alter quantitative metrics when applied over the entire dataset.
    
    \item \textit{In-context learning}~\cite{lampinen2022can} is a type of discrete, few-shot prompt. We begin with the null prompt and prepend one, two, or four task examples using the same template. Particular examples are chosen by computing the k-nearest neighbors~\cite{johnson2019billion} of the findings section for each training example in the embedding space of a PubMedBERT model~\cite{deka2022evidence}. This provides the most relevant examples for the RRS task.
    
    \item \textit{Prefix tuning}~\cite{li2021prefix} is a parameter-efficient fine-tuning method which prepends and optimizes an additional task-specific vector called the \textit{prefix} as input to the model. For our base and large architectures, this requires tuning a mere 0.17\% and 0.13\% of total parameters, respectively (see Table~\ref{tab:num-params}). This approach provides the model a task-specific prompt that is very well aligned to the downstream task.
    
    \item \textit{LoRA}~\cite{hu2021lora}, or low-rank adaptation, approximates the fine-tuning process by injecting trainable rank decomposition matrices into each architecture layer. Compared to prefix tuning, which requires some portion of the input sequence for adaptation, LoRA beneficially preserves the entire sequence length for the downstream task. For our base and large architectures, this requires tuning a mere 0.39\% and 0.32\% of total parameters, respectively (see Table~\ref{tab:num-params}). Because LoRA modifies slightly more parameters than prefix tuning, we characterize this method as having greater domain adaptation.
\end{enumerate} 
\section{Experiments}

\subsection{Data}
\label{label:data}

\begin{table}[b!]
\centering
\caption{Number of reports in MIMIC-III by modality, anatomy, and dataset split.}

\resizebox{0.4\textwidth}{!}{
\begin{tabular}{l|ccc}
\hline
\textbf{Modality/} & \multicolumn{3}{c}{\textbf{Number of reports}} \\
\textbf{Anatomy} & \textbf{Train} & \textbf{Val} & \textbf{Test} \\
\hline
CT head & 25,122 & 3,140 & 3,141 \\
CT abdomen & 12,792 & 1,599 & 1,599 \\
CT chest & 10,229 & 1,278 & 1,280 \\
MR head & 5,851 & 731 & 732 \\
CT spine & 4,414 & 551 & 553 \\
CT neck & 912 & 114 & 115 \\
MR spine & - & - & 2,822 \\
CT sinus & - & - & 1,268 \\
MR abdomen & - & - & 1,062 \\
MR pelvis & - & - & 254 \\
MR neck & - & - & 231 \\
\hline
\end{tabular}
}
\vspace{0.2cm}
\label{table:dataset}
\end{table}
\begin{figure*}[htp]
    \centering
    \includegraphics[width = 1\textwidth]{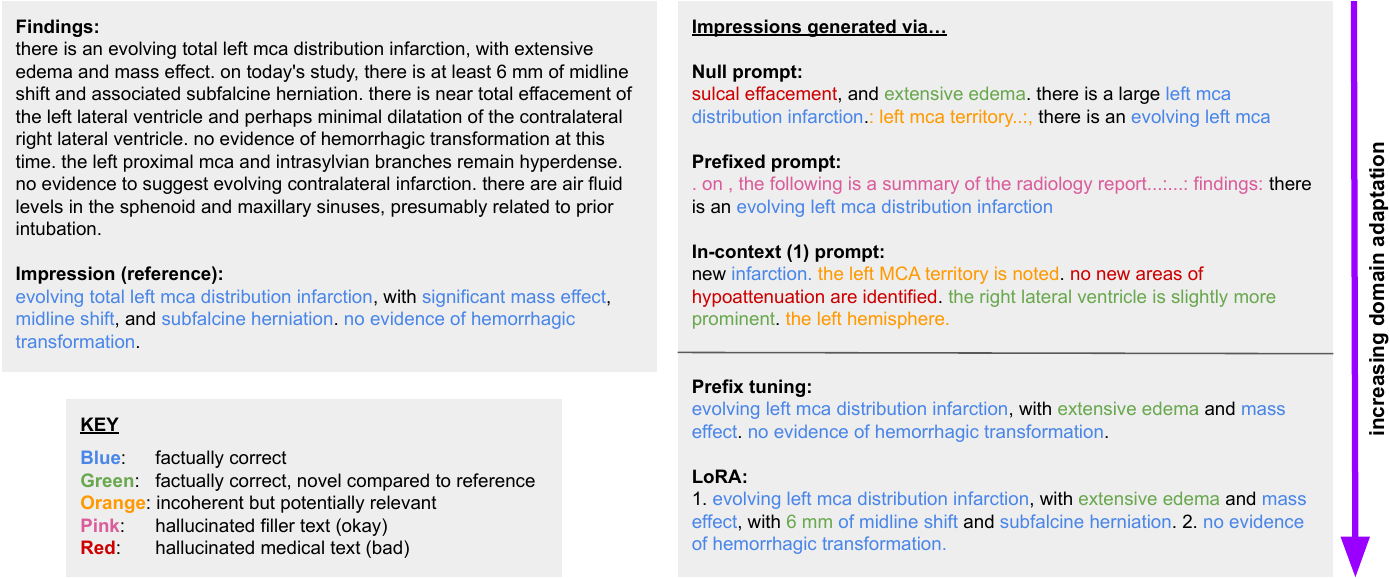}
    \caption{Example radiology report. \underline{Left}: Findings and reference impression. \underline{\smash{Right}}: Generated impressions with various methods for discrete prompting (top) and parameter-efficient fine-tuning (bottom), all using the \clintfivel~model. Color annotations were provided by a radiologist who specializes in the relevant anatomy (head).}
    \label{fig:example_output}
\end{figure*}

Our primary dataset is MIMIC-III~\cite{johnson2016mimic}, which contains 79,790 radiology reports across two imaging modalities and seven anatomies.  Table~\ref{table:dataset} contains a dataset overview. Figure~\ref{fig:example_output} contains an example of a CT head report; recall the task of RRS is to predict the report impressions section (label) given the findings section (input).
We provide secondary evaluations on another radiology report dataset, MIMIC-CXR~\cite{johnson2019mimiccxr}. This is an easier summarization task as it contains one modality and anatomy (chest x-rays) with generally shorter impression sections than MIMIC-III.
PhysioNet~\cite{johnson2020physionet} and ViLMedic~\cite{delbrouck2022vilmedic} provided access to pre-processed versions of these datasets, removing confidential patient information. 
Lastly, to demonstrate results beyond the MIMIC suite, we incorporate a supplemental dataset of 10,721 reports from ultrasound abdomen exams acquired at Stanford Hospital. This dataset was acquired and analyzed with institutional review board approval.

\subsection{Evaluation}
\label{label:eval}
For quantitative evaluation of our generated impressions, we employ common summarization metrics such as BLEU and ROUGE-L. Between a given pair of reference and generated text, BLEU evaluates overlap using a weighted average of 1- to 4-gram precision, and ROUGE-L evaluates the longest common subsequence overlap. Beyond these token-level syntactic similarity metrics, we employ by also using metrics like BERTScore (via HuggingFace), which computes the semantic similarity between the reference and generated texts using BERT embeddings~\cite{zhang2019bertscore}. Lastly, following previous work~\cite{delbrouck2022radgraph}, we evaluate our model with F1-RadGraph, a F-score style metric that measures the factual correctness, consistency and completeness of generated radiology reports compared to the reference. F1-RadGraph uses RadGraph~\cite{jain2021radgraph}, a graph dataset of entities and relations present in radiology reports.

\begin{figure}[b!]
\centering
  \includegraphics[width=0.48\textwidth]{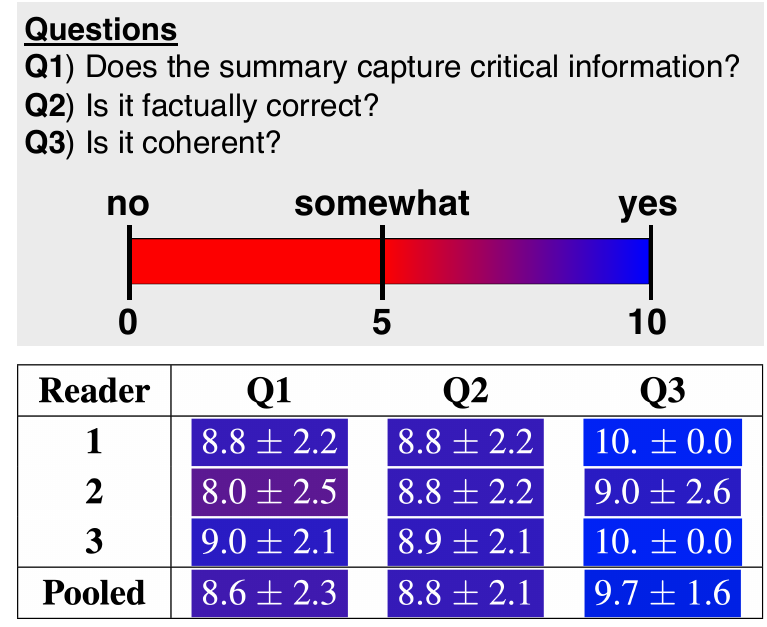}
  \caption{Radiology reader study. \underline{\smash{Top}}: Study design. \underline{\smash{Bottom}}: Results via \clintfivel~+ LoRA on random samples from the CT head dataset. The model scores highest in coherence (Q3) and generally performs well capturing critical information (Q1) in a factually correct way (Q2). Each entry's highlight color corresponds to its location on the above color spectrum.}
  \label{fig:reader_study}
\end{figure}
\begin{figure*}[t]
    \centering
    \includegraphics[width = 1\textwidth]{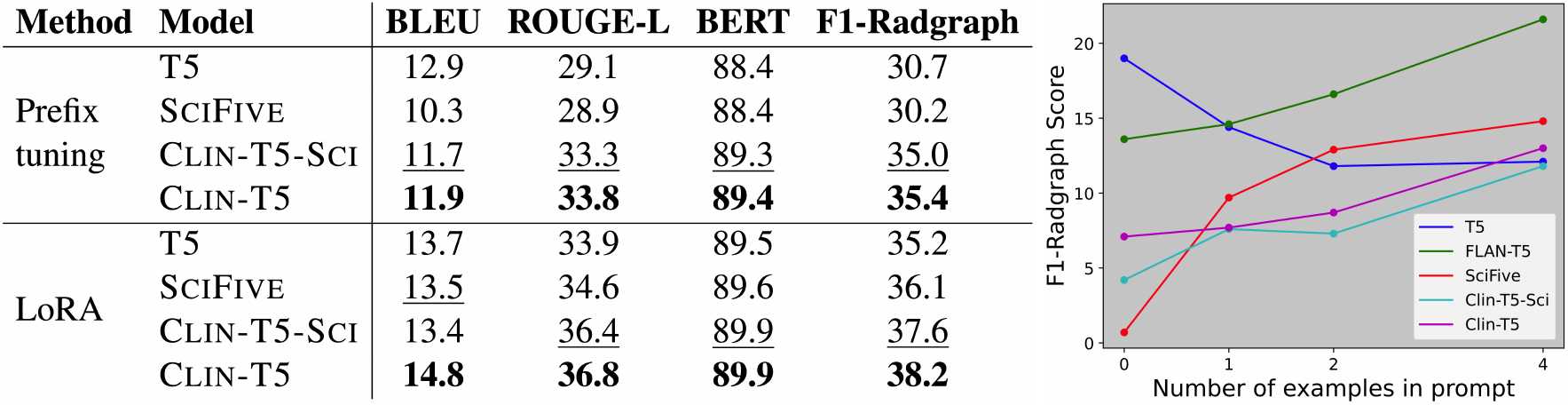}
    \caption{Domain adaptation. \underline{\smash{Left}}: Adaptation via pretraining on increasingly relevant data (\tfive, \scifive, \clintfivesci, \clintfive) generally leads to improved performance for both fine-tuning methods.
    Note we exclude \flantfive, whose degree of domain adaptation is difficult to rank. See Table~\ref{tab:results-base} in the appendix for comprehensive results. \underline{\smash{Right}}: Adaptation via increasing number of in-context examples leads to improved performance in most models. See Section~\ref{sec:results-domain-adaptation} for discussion.}
    \label{fig:domain-adapt}
\end{figure*}

For qualitative evaluation, we include a reader study (Figure~\ref{fig:reader_study}) with three board-certified radiologists. Each evaluated a randomly selected group of twenty generated impressions in comparison to  reference impressions. They responded with either 0 (``no''), 5 (``somewhat''), or 10 (``yes'') to the following three questions:
\begin{enumerate}
    \item Does the generated impression capture critical information? Consider if the model missed any important findings.
    \item Is the generated impression factually correct? Given the text that the model did generate, consider its correctness.
    \item Is the generated impression coherent, i.e.~do you find the syntax comprehensible?
\end{enumerate}

\subsection{Experimental details}\label{sec:experiment-details}
No further model tuning was needed for the null, prefix, and in-context discrete prompting experiments; we simply run inference with the pre-existing LLM. When tuning model hyperparameters for prefix tuning and LoRA, we discovered that the same set of hyperparameters achieves best performance across all five models. This seems reasonable given that each model employs the same architecture. As such we tuned each model with the same set of hyperparameters. Please see Section~\ref{sec:hyperparam_appendix} in the appendix for details.

\section{Results and Discussion}

\subsection{Domain Adaptation}\label{sec:results-domain-adaptation}
A recurring theme throughout our results is that increased domain adaptation leads to improved performance. One axis for domain adaptation is the relevance of pretraining data to the evaluation task, in our case RRS on the MIMIC-III dataset. 
Figure~\ref{fig:domain-adapt} (left) demonstrates that pretraining the same architecture on increasingly relevant data (\tfive, \scifive, \clintfivesci, \clintfive) typically improves performance for prefix tuning and LoRA. We do not include \flantfive~in this portion of the analysis, as its relative degree of domain adaptation is difficult to assess.

The second axis for domain adaptation is the amount of context provided either via longer discrete prompts (null, prefix, in-context prompting), or parameter-efficient fine-tuning (prefix tuning, LoRA). For all models, fine-tuning significantly outperforms any discrete prompting technique. 
Discrete prompting is still useful, however, especially if one only has black-box access to the LLM. Therefore, we compare techniques for discrete prompting in Figure~\ref{fig:domain-adapt} (right), demonstrating that prompting the model with more in-context examples improves performance for almost every model with the lone exception of \tfive. 
As \tfive~is the least domain-adapted model, perhaps a small number of in-context examples may actually hurt performance when examples are sufficiently out-of-domain. We leave this question to future work.
Another component of Figure~\ref{fig:domain-adapt} (right) is that the remaining four models improve at different rates: \scifive~seems to level off near four in-context examples, while the others continue to steadily increase. This supports our hypothesis that instruction prompt tuning and maximal domain adaptation (in this case, to clinical text) improve results seen with in-context learning. This also motivates future work in instruction prompt tuning a model using domain-specific biomedical or clinical instructions.

\setlength{\tabcolsep}{6pt} 

\begin{table*}[t]
\caption{Best results overall. \underline{\smash{Top}}: Given that the base architecture (223M parameters) performs best via pretraining on clinical text (\clintfive) and subsequent fine-tuning, we improve performance on MIMIC-III by scaling to the large architecture (738M). \underline{Bottom}: LoRA also outperforms prefix tuning on the MIMIC-CXR dataset using \clintfive.
}
\resizebox{1 \textwidth}{!}{
\centering
\begin{tabular}{l l l | c c c c c}
\textbf{Dataset} & \textbf{Method} & \textbf{Size} & \textbf{BLEU} & \textbf{ROUGE-L} & \textbf{BERT} & \textbf{F1-Radgraph} & \textbf{F1-CheXbert} \\
\hline
\multirow{4}{*}{MIMIC-III} & \multirow{2}{*}{prefix tuning}  & base & 11.9 & 33.8 & 89.4 & 35.4 & - \\
&  & large & \underline{14.6} & \underline{36.7} & \underline{89.9} & \underline{38.4} & -  \\
& \multirow{2}{*}{LoRA} & base & 14.5 & 36.4 & 89.9 & 38.0 & -  \\
& & large & \textbf{16.2} & \textbf{38.7} & \textbf{90.2} & \textbf{40.8} & -  \\
\hline
\multirow{2}{*}{MIMIC-CXR} & prefix tuning & large & 16.1 & 43.4 & 89.7 & 41.0 & 70.2  \\
 & LoRA & large & \textbf{18.9} & \textbf{44.5} & \textbf{90.0} & \textbf{41.8} & \textbf{70.9} \\

\end{tabular}
}
\label{tab:best}
\end{table*}
\setlength{\tabcolsep}{2.1pt} 

\begin{table*}[b!]
\caption{Out-of-distribution (OOD) performance of \clintfive~prefix tuned on CT head. Compared to in-distribution (first row), performance suffers increasingly with OOD modalities (second row) and anatomies (third row). Additionally, when evaluating CT head, tuning on a larger dataset comprising all modalities/anatomies (bottom row) improves performance compared to tuning on CT head alone (top row).
}
\centering
\begin{tabular}{c c | c c | c c c c}
 \multicolumn{2}{c}{\textbf{\underline{Dataset}}} & \multicolumn{2}{c}{\textbf{\underline{OOD}}} & & & & \\
\textbf{Train} & \textbf{Test} & \textbf{Modality} & \textbf{Anatomy} & \textbf{BLEU} & \textbf{ROUGE-L} & \textbf{BERT} & \textbf{F1-Radgraph} \\

\hline
CT head & CT head & & & \underline{11.4} & \underline{35.0} & \textbf{89.8} & \underline{35.1} \\
CT head & MR head & \ding{51} & & 9.0 & 27.5 & 87.8 & 27.4 \\
CT head & CT other & & \ding{51} & 2.9 & 19.5 & 86.7 & 16.3 \\
CT head & MR other & \ding{51} & \ding{51} & 7.9 & 24.2 & 87.2 & 25.9 \\
All & CT head & N/A & N/A & \textbf{12.6} & \textbf{35.3} & \underline{89.7} & \textbf{36.4} \\
\end{tabular}
\label{tab:ood}
\end{table*}

Table~\ref{tab:results-base} in the appendix includes an ablation of all configurations for domain adaptation using a base architecture (223M parameters). We subsequently take the best model (\clintfive) for the best methods (prefix tuning, LoRA) and scale to a large architecture (738M parameters) in Table~\ref{tab:best}. This scaling provides a significant performance boost. Similar to the base architecture, LoRA outperforms prefix tuning in all cases. Interestingly, \clintfiveb~+ LoRA achieves nearly equivalent performance to \clintfivel~+ prefix tuning, exemplifying the benefits of LoRA over prefix tuning. Finally, after performing all prior analysis using the MIMIC-III dataset, we apply the best combination (\clintfivel~+ LoRA) on MIMIC-CXR and include results in Table~\ref{tab:best}.

\subsection{Out-of-distribution performance}\label{sec:results-ood}
Table~\ref{tab:ood} demonstrates out-of-distribution performance using a \clintfiveb~model prefix tuned on CT head data. When evaluating on a different test set, the model better summarizes reports on a different modality (MR head) than a different anatomy (CT other). This suggests that report finding tokens are more anatomy- than modality-specific, which seems reasonable. Counterintuitively, the model performs worse when shifting just the anatomy (CT other) compared to shifting the modality and anatomy (MR other). This could be due to a myriad of reasons, such as MR findings being 15\% longer than CT findings and MR other containing fewer anatomies (four) than CT other (five). Lastly, we find that training \clintfiveb~on all data leads to higher performance than training on CT head alone. 

\subsection{Qualitative Evaluation}

\subsubsection{Error Analysis}\label{sec:result-error-analysis}
We perform a qualitative error analysis of 20 randomly selected reference findings, reference impressions, and generated impressions via \clintfivel~on the CT head dataset. Please see Figure~\ref{fig:example_output} for an example. We describe four types of deviations from the reference impressions and their corresponding colors used in Figure~\ref{fig:example_output}:
\begin{enumerate}
    \item Factually correct text that is novel compared to the reference impression (green). This text is present in the reference findings but not in the reference impression.
    \item Incoherent but potentially relevant text (orange). This contains medically relevant information that is also included in the reference, but is presented with incoherent grammar.
    \item Hallucinated filler text (pink). This includes extra punctuation or common words such as ``findings.'' These are filler text because they are undesirable but do not detract from the correctness of the generated impression.
    \item Hallucinated medical text (red). This includes text that is either (1) not explicitly included in the findings or reference impression, or (2) relevant to the findings but communicated in a factually incorrect manner. This is the worst of the four deviations, as we want the model to avoid inferring information which didn't originate directly from the radiologist.
\end{enumerate}

\subsubsection{Reader study}\label{sec:results-reader-study}
Three radiologists evaluated impressions generated via \clintfivel~+ LoRA on the CT head dataset according to the procedure described in Section~\ref{label:eval}. Results in Figure~\ref{fig:reader_study} are encouraging, as our generated impressions scored well for all three questions. Additionally, the radiologists shared the following observations:

\begin{itemize}
    \item Occasionally there is extra information included in the ``reference'' impression that is not available in the findings, i.e.~which the model has no chance of summarizing.
    \item The model may include duplicate or re-summarized when referring to prior studies. For example, the reference will state ``area of subarachnoid hemorrhage ... which is unchanged since the patient's prior scan,'' while the generated impression merely says ``no significant change since the prior study.'' This difference is typically an institutional or personal preference.
    \item The model made an incorrect reference to the patient's prior medical history. The reference was ``subtle hypodensity in the left frontal lobe. given lack of prior studies available for comparison,'' but the generated impression was ``subtle hypodensity ... in the left frontal lobe, which is consistent with prior studies.''
\end{itemize}

This study provided valuable insights which could not be obtained via quantitative metrics. Fundamentally we advocate for the use of reader studies when evaluating report summarization to facilitate more clinically relevant research.

\subsection{Pitfalls}\label{sec:results-pitfalls}

We now discuss weaknesses in our analysis which motivate future study. One potential concern is that \clintfive~and \clintfivesci~were pretrained over the MIMIC-III dataset. This serves our purpose for evaluating models with high levels of domain adaptation, but it could result in data leakage if~\citet{lehman2023clint5} pretrained on the test set. Hence we evaluate on a separate dataset of ultrasound exams and observe that \clintfive~and \clintfivesci~similarly perform the best (Table~\ref{tab:results-suh-lora}, appendix). This bolsters evidence that adaptation via pretraining on clinical text benefits RRS.

Another weakness is that ``domain'' and ``distribution'' are not rigorously defined. Our intuitive characterizations could be improved by quantifying or visualizing the distance between different distributions using various embedding- or graph-based methods~\cite{johnson2019billion, jain2021radgraph}.

Lastly, we made assumptions determining the best configuration of model and prompting method. We first performed a comprehensive evaluation of all models and methods using the base architecture (223M parameters) on MIMIC-III (Table~\ref{tab:results-base}, appendix) and all models with LoRA on Stanford's ultrasound dataset (Table~\ref{tab:results-suh-lora}, appendix). From these results we chose only the best model (\clintfive) for scaling to the large architecture (738M parameters) due to compute constraints. Future work should include a comprehensive evaluation of all configurations across datasets and architecture sizes.
\vspace{1.65mm}
\vspace{-4mm}
\section{Conclusion}

Our research employs innovative lightweight strategies to adapt LLMs for the task of RRS. We investigate how domain adaptation---both via model pretraining on relevant data and via methods for discrete prompting and parameter-efficient fine-tuning---affects downstream RRS task performance. We achieve best performance using a model pretrained on clinical text (\clintfive) and subsequently fine-tuned with RRS samples using LoRA. These compelling results require tuning a mere 0.32\% of model parameters. While further validation is required before clinical deployment, we believe our findings contribute to the literature and advance the potential for improved radiologist workflows and patient care.
\vspace{1.65mm}
\vspace{-4mm}
\section{Acknowledgements}

This work is a continuation of the radiology report summarization track at ACL BioNLP~\cite{DelbrouckRadSum23}. We received support from NIH contracts 75N92020C00008 and 75N92020C00021.

\clearpage
\bibliography{refs}

\begin{thebibliography}{42}
\expandafter\ifx\csname natexlab\endcsname\relax\def\natexlab#1{#1}\fi

\bibitem[{Abacha et~al.(2021)Abacha, M’rabet, Zhang, Shivade, Langlotz, and
  Demner-Fushman}]{abacha2021overview}
Asma~Ben Abacha, Yassine M’rabet, Yuhao Zhang, Chaitanya Shivade, Curtis
  Langlotz, and Dina Demner-Fushman. 2021.
\newblock Overview of the mediqa 2021 shared task on summarization in the
  medical domain.
\newblock In \emph{Proceedings of the 20th Workshop on Biomedical Language
  Processing}, pages 74--85.

\bibitem[{Brown et~al.(2020)Brown, Mann, Ryder, Subbiah, Kaplan, Dhariwal,
  Neelakantan, Shyam, Sastry, Askell et~al.}]{brown2020language}
Tom Brown, Benjamin Mann, Nick Ryder, Melanie Subbiah, Jared~D Kaplan, Prafulla
  Dhariwal, Arvind Neelakantan, Pranav Shyam, Girish Sastry, Amanda Askell,
  et~al. 2020.
\newblock Language models are few-shot learners.
\newblock \emph{Advances in neural information processing systems},
  33:1877--1901.

\bibitem[{Chowdhery et~al.(2022)Chowdhery, Narang, Devlin, Bosma, Mishra,
  Roberts, Barham, Chung, Sutton, Gehrmann et~al.}]{chowdhery2022palm}
Aakanksha Chowdhery, Sharan Narang, Jacob Devlin, Maarten Bosma, Gaurav Mishra,
  Adam Roberts, Paul Barham, Hyung~Won Chung, Charles Sutton, Sebastian
  Gehrmann, et~al. 2022.
\newblock Palm: Scaling language modeling with pathways.
\newblock \emph{arXiv preprint arXiv:2204.02311}.

\bibitem[{Chung et~al.(2022{\natexlab{a}})Chung, Hou, Longpre
  et~al.}]{chung2022flant5}
H.W. Chung, L.~Hou, S.~Longpre, et~al. 2022{\natexlab{a}}.
\newblock Scaling instruction-finetuned language models.
\newblock \emph{https://doi.org/10.48550/arXiv.2210.11416}.

\bibitem[{Chung et~al.(2022{\natexlab{b}})Chung, Hou, Longpre, Zoph, Tay,
  Fedus, Li, Wang, Dehghani, Brahma et~al.}]{chung2022scaling}
Hyung~Won Chung, Le~Hou, Shayne Longpre, Barret Zoph, Yi~Tay, William Fedus,
  Eric Li, Xuezhi Wang, Mostafa Dehghani, Siddhartha Brahma, et~al.
  2022{\natexlab{b}}.
\newblock Scaling instruction-finetuned language models.
\newblock \emph{arXiv preprint arXiv:2210.11416}.

\bibitem[{Dai et~al.(2021)Dai, Wang, Lyu, and Zhu}]{dai2021bdkg}
Songtai Dai, Quan Wang, Yajuan Lyu, and Yong Zhu. 2021.
\newblock Bdkg at mediqa 2021: system report for the radiology report
  summarization task.
\newblock In \emph{Proceedings of the 20th Workshop on Biomedical Language
  Processing}, pages 103--111.

\bibitem[{Deka et~al.(2022)Deka, Jurek-Loughrey et~al.}]{deka2022evidence}
Pritam Deka, Anna Jurek-Loughrey, et~al. 2022.
\newblock Evidence extraction to validate medical claims in fake news
  detection.
\newblock In \emph{International Conference on Health Information Science},
  pages 3--15. Springer.

\bibitem[{Delbrouck et~al.(2022{\natexlab{a}})Delbrouck, Chambon, Bluethgen,
  Tsai, Almusa, and Langlotz}]{delbrouck2022radgraph}
Jean-Benoit Delbrouck, Pierre Chambon, Christian Bluethgen, Emily Tsai, Omar
  Almusa, and Curtis Langlotz. 2022{\natexlab{a}}.
\newblock Improving the factual correctness of radiology report generation with
  semantic rewards.
\newblock \emph{https://aclanthology.org/2022.findings-emnlp.319}.

\bibitem[{Delbrouck et~al.(2022{\natexlab{b}})Delbrouck, Saab, Varma, Eyuboglu,
  Chambon, Dunnmon, Zambrano, Chaudhari, and Langlotz}]{delbrouck2022vilmedic}
Jean-benoit Delbrouck, Khaled Saab, Maya Varma, Sabri Eyuboglu, Pierre Chambon,
  Jared Dunnmon, Juan Zambrano, Akshay Chaudhari, and Curtis Langlotz.
  2022{\natexlab{b}}.
\newblock \href {https://doi.org/10.18653/v1/2022.acl-demo.3} {{V}i{LM}edic: a
  framework for research at the intersection of vision and language in medical
  {AI}}.
\newblock In \emph{Proceedings of the 60th Annual Meeting of the Association
  for Computational Linguistics: System Demonstrations}, pages 23--34, Dublin,
  Ireland. Association for Computational Linguistics.

\bibitem[{Delbrouck et~al.(2023)Delbrouck, Varma, Chambon, and
  Langlotz}]{DelbrouckRadSum23}
Jean-Benoit Delbrouck, Maya Varma, Pierre Chambon, and Curtis Langlotz. 2023.
\newblock Overview of the radsum23 shared task on multi-modal and
  multi-anatomical radiology report summarization.
\newblock In \emph{Proceedings of the 22st Workshop on Biomedical Language
  Processing}, Toronto, Canada. Association for Computational Linguistics.

\bibitem[{Devlin et~al.(2018)Devlin, Chang, Lee, and
  Toutanova}]{devlin2018bert}
Jacob Devlin, Ming-Wei Chang, Kenton Lee, and Kristina Toutanova. 2018.
\newblock Bert: Pre-training of deep bidirectional transformers for language
  understanding.
\newblock \emph{arXiv preprint arXiv:1810.04805}.

\bibitem[{Gershanik et~al.(2011)Gershanik, Lacson, and
  Khorasani}]{gershanik2011critical}
Esteban~F Gershanik, Ronilda Lacson, and Ramin Khorasani. 2011.
\newblock Critical finding capture in the impression section of radiology
  reports.
\newblock In \emph{AMIA Annual Symposium Proceedings}, volume 2011, page 465.
  American Medical Informatics Association.

\bibitem[{Houlsby et~al.(2019)Houlsby, Giurgiu, Jastrzebski, Morrone,
  De~Laroussilhe, Gesmundo, Attariyan, and Gelly}]{houlsby2019parameter}
Neil Houlsby, Andrei Giurgiu, Stanislaw Jastrzebski, Bruna Morrone, Quentin
  De~Laroussilhe, Andrea Gesmundo, Mona Attariyan, and Sylvain Gelly. 2019.
\newblock Parameter-efficient transfer learning for nlp.
\newblock In \emph{International Conference on Machine Learning}, pages
  2790--2799. PMLR.

\bibitem[{Hu et~al.(2021)Hu, Shen, Wallis, Allen-Zhu, Li, Wang, and
  Chen}]{hu2021lora}
Edward Hu, Yelong Shen, Phil Wallis, Zeyuan Allen-Zhu, Yuanzhi Li, Lu~Wang, and
  Weizhu Chen. 2021.
\newblock \href {http://arxiv.org/abs/2106.09685} {Lora: Low-rank adaptation of
  large language models}.

\bibitem[{Irvin et~al.(2019)}]{irvin2019chexpert}
Jeremy Irvin et~al. 2019.
\newblock Chexpert: A large chest radiograph dataset with uncertainty labels
  and expert comparison.
\newblock \emph{https://doi.org/10.48550/arXiv.1901.07031}.

\bibitem[{Jain et~al.(2021)Jain, Agrawal, Saporta, Truong, Duong, Bui, Chambon,
  Zhang, Lungren, Ng et~al.}]{jain2021radgraph}
Saahil Jain, Ashwin Agrawal, Adriel Saporta, Steven~QH Truong, Du~Nguyen Duong,
  Tan Bui, Pierre Chambon, Yuhao Zhang, Matthew~P Lungren, Andrew~Y Ng, et~al.
  2021.
\newblock Radgraph: Extracting clinical entities and relations from radiology
  reports.
\newblock \emph{arXiv preprint arXiv:2106.14463}.

\bibitem[{Johnson et~al.(2020{\natexlab{a}})Johnson, Bulgarelli, Pollard,
  Horng, Celi, and Mark}]{johnson2020mimic}
Alistair Johnson, Lucas Bulgarelli, Tom Pollard, Steven Horng, Leo~Anthony
  Celi, and Roger Mark. 2020{\natexlab{a}}.
\newblock Mimic-iv.
\newblock \emph{PhysioNet. Available online at: https://physionet.
  org/content/mimiciv/1.0/(accessed August 23, 2021)}.

\bibitem[{Johnson et~al.(2020{\natexlab{b}})Johnson, Pollard, and
  Mark}]{johnson2020physionet}
Alistair Johnson, Tom Pollard, and Roger Mark. 2020{\natexlab{b}}.
\newblock {MIMIC-III} clinical database.

\bibitem[{Johnson et~al.(2019{\natexlab{a}})}]{johnson2019mimiccxr}
Alistair Johnson et~al. 2019{\natexlab{a}}.
\newblock Mimic-cxr, a de-identified publicly available database of chest
  radiographs with free-text reports.
\newblock \emph{https://www.nature.com/articles/s41597-019-0322-0}.

\bibitem[{Johnson et~al.(2016)Johnson, Pollard, Shen, Lehman, Feng, Ghassemi,
  Moody, Szolovits, Anthony~Celi, and Mark}]{johnson2016mimic}
Alistair~EW Johnson, Tom~J Pollard, Lu~Shen, Li-wei~H Lehman, Mengling Feng,
  Mohammad Ghassemi, Benjamin Moody, Peter Szolovits, Leo Anthony~Celi, and
  Roger~G Mark. 2016.
\newblock Mimic-iii, a freely accessible critical care database.
\newblock \emph{Scientific data}, 3(1):1--9.

\bibitem[{Johnson et~al.(2019{\natexlab{b}})Johnson, Douze, and
  J{\'e}gou}]{johnson2019billion}
Jeff Johnson, Matthijs Douze, and Herv{\'e} J{\'e}gou. 2019{\natexlab{b}}.
\newblock Billion-scale similarity search with {GPUs}.
\newblock \emph{IEEE Transactions on Big Data}, 7(3):535--547.

\bibitem[{Kahn~Jr et~al.(2009)Kahn~Jr, Langlotz, Burnside, Carrino, Channin,
  Hovsepian, and Rubin}]{kahn2009toward}
Charles~E Kahn~Jr, Curtis~P Langlotz, Elizabeth~S Burnside, John~A Carrino,
  David~S Channin, David~M Hovsepian, and Daniel~L Rubin. 2009.
\newblock Toward best practices in radiology reporting.
\newblock \emph{Radiology}, 252(3):852--856.

\bibitem[{Lampinen et~al.(2022)Lampinen, Dasgupta, Chan, Matthewson, Tessler,
  Creswell, McClelland, Wang, and Hill}]{lampinen2022can}
Andrew~K Lampinen, Ishita Dasgupta, Stephanie~CY Chan, Kory Matthewson,
  Michael~Henry Tessler, Antonia Creswell, James~L McClelland, Jane~X Wang, and
  Felix Hill. 2022.
\newblock Can language models learn from explanations in context?
\newblock \emph{arXiv preprint arXiv:2204.02329}.

\bibitem[{Lehman and Johnson(2023)}]{lehman2023clint5}
E.~Lehman and A.~Johnson. 2023.
\newblock Clinical-t5: Large language models built using mimic clinical text.
\newblock \emph{https://doi.org/10.13026/rj8x-v335}.

\bibitem[{Lester et~al.(2021)Lester, Al-Rfou, and Constant}]{lester2021power}
Brian Lester, Rami Al-Rfou, and Noah Constant. 2021.
\newblock The power of scale for parameter-efficient prompt tuning.
\newblock \emph{arXiv preprint arXiv:2104.08691}.

\bibitem[{Li and Liang(2021)}]{li2021prefix}
Xiang~Lisa Li and Percy Liang. 2021.
\newblock Prefix-tuning: Optimizing continuous prompts for generation.
\newblock \emph{arXiv preprint arXiv:2101.00190}.

\bibitem[{Lin et~al.(2020)Lin, Madotto, and Fung}]{lin2020exploring}
Zhaojiang Lin, Andrea Madotto, and Pascale Fung. 2020.
\newblock Exploring versatile generative language model via parameter-efficient
  transfer learning.
\newblock \emph{arXiv preprint arXiv:2004.03829}.

\bibitem[{Liu et~al.(2023)Liu, Yuan, Fu, Jiang, Hayashi, and
  Neubig}]{liu2023prompt}
Pengfei Liu, Weizhe Yuan, Jinlan Fu, Zhengbao Jiang, Hiroaki Hayashi, and
  Graham Neubig. 2023.
\newblock \href {https://doi.org/10.1145/3560815} {Pre-train, prompt, and
  predict: A systematic survey of prompting methods in natural language
  processing}.
\newblock \emph{ACM Comput. Surv.}, 55(9).

\bibitem[{Longpre et~al.(2023)Longpre, Hou, Vu, Webson, Chung, Tay, Zhou, Le,
  Zoph, Wei, and Roberts}]{longpre2023flan}
Shayne Longpre, Le~Hou, Tu~Vu, Albert Webson, Hyung~Won Chung, Yi~Tay, Denny
  Zhou, Quoc~V. Le, Barret Zoph, Jason Wei, and Adam Roberts. 2023.
\newblock \href {http://arxiv.org/abs/2301.13688} {The flan collection:
  Designing data and methods for effective instruction tuning}.

\bibitem[{Miura et~al.(2021)Miura, Zhang, Tsai, Langlotz, and
  Jurafsky}]{miura2021}
Yasuhide Miura, Yuhao Zhang, Emily~Bao Tsai, Curtis~P. Langlotz, and Dan
  Jurafsky. 2021.
\newblock Improving factual completeness and consistency of image-to-text
  radiology report generation.
\newblock In \emph{NAACL-HLT 2021}.

\bibitem[{NCBI(1996)}]{pubmed}
NCBI. 1996.
\newblock \href {https://pubmed.ncbi.nlm.nih.gov/} {Pubmed}.

\bibitem[{NCBI(2000)}]{pmc}
NCBI. 2000.
\newblock \href {https://www.ncbi.nlm.nih.gov/pmc} {Pubmed central (pmc)}.

\bibitem[{Phan et~al.(2021)Phan, Anibal, Tran, Chanana, Bahadroglu, Peltekian,
  and Altan-Bonnet}]{phan2021scifive}
Long~N Phan, James~T Anibal, Hieu Tran, Shaurya Chanana, Erol Bahadroglu, Alec
  Peltekian, and Gr{\'e}goire Altan-Bonnet. 2021.
\newblock Scifive: a text-to-text transformer model for biomedical literature.
\newblock \emph{arXiv preprint arXiv:2106.03598}.

\bibitem[{Radford et~al.(2019)Radford, Wu, Child, Luan, Amodei, Sutskever
  et~al.}]{radford2019language}
Alec Radford, Jeffrey Wu, Rewon Child, David Luan, Dario Amodei, Ilya
  Sutskever, et~al. 2019.
\newblock Language models are unsupervised multitask learners.
\newblock \emph{OpenAI blog}, 1(8):9.

\bibitem[{Raffel et~al.(2020)Raffel, Shazeer, Roberts, Lee, Narang, Matena,
  Zhou, Li, and Liu}]{raffel2020t5}
Colin Raffel, Noam Shazeer, Adam Roberts, Katherine Lee, Sharan Narang, Michael
  Matena, Yanqi Zhou, Wei Li, and Peter~J Liu. 2020.
\newblock Exploring the limits of transfer learning with a unified text-to-text
  transformer.
\newblock \emph{The Journal of Machine Learning Research}, 21(1):5485--5551.

\bibitem[{Rebuffi et~al.(2017)Rebuffi, Bilen, and
  Vedaldi}]{rebuffi2017learning}
Sylvestre-Alvise Rebuffi, Hakan Bilen, and Andrea Vedaldi. 2017.
\newblock Learning multiple visual domains with residual adapters.
\newblock \emph{Advances in neural information processing systems}, 30.

\bibitem[{Singhal et~al.(2022)Singhal, Azizi, Tu, Mahdavi, Wei, Chung, Scales,
  Tanwani, Cole-Lewis, Pfohl et~al.}]{singhal2022large}
Karan Singhal, Shekoofeh Azizi, Tao Tu, S~Sara Mahdavi, Jason Wei, Hyung~Won
  Chung, Nathan Scales, Ajay Tanwani, Heather Cole-Lewis, Stephen Pfohl, et~al.
  2022.
\newblock Large language models encode clinical knowledge.
\newblock \emph{arXiv preprint arXiv:2212.13138}.

\bibitem[{Vaswani et~al.(2017)Vaswani, Shazeer, Parmar, Uszkoreit, Jones,
  Gomez, Kaiser, and Polosukhin}]{vaswani2017attention}
Ashish Vaswani, Noam Shazeer, Niki Parmar, Jakob Uszkoreit, Llion Jones,
  Aidan~N Gomez, {\L}ukasz Kaiser, and Illia Polosukhin. 2017.
\newblock Attention is all you need.
\newblock \emph{Advances in neural information processing systems}, 30.

\bibitem[{Wei et~al.(2022)Wei, Wang, Schuurmans, Bosma, Chi, Le, and
  Zhou}]{wei2022chain}
Jason Wei, Xuezhi Wang, Dale Schuurmans, Maarten Bosma, Ed~Chi, Quoc Le, and
  Denny Zhou. 2022.
\newblock Chain of thought prompting elicits reasoning in large language
  models.
\newblock \emph{arXiv preprint arXiv:2201.11903}.

\bibitem[{Zhang* et~al.(2020)Zhang*, Kishore*, Wu*, Weinberger, and
  Artzi}]{zhang2019bertscore}
Tianyi Zhang*, Varsha Kishore*, Felix Wu*, Kilian~Q. Weinberger, and Yoav
  Artzi. 2020.
\newblock \href {https://openreview.net/forum?id=SkeHuCVFDr} {Bertscore:
  Evaluating text generation with bert}.
\newblock In \emph{International Conference on Learning Representations}.

\bibitem[{Zhang et~al.(2020)Zhang, Merck, Tsai, Manning, and
  Langlotz}]{zhang2020}
Yuhao Zhang, Derek Merck, Emily~Bao Tsai, Christopher~D. Manning, and Curtis~P.
  Langlotz. 2020.
\newblock Optimizing the factual correctness of a summary: A study of
  summarizing radiology reports.
\newblock In \emph{ACL2020}.

\bibitem[{Zhao and Sch{\"u}tze(2021)}]{zhao2021discrete}
Mengjie Zhao and Hinrich Sch{\"u}tze. 2021.
\newblock Discrete and soft prompting for multilingual models.
\newblock \emph{arXiv preprint arXiv:2109.03630}.

\end{thebibliography}
\bibliographystyle{acl_natbib}

\clearpage
\appendix
\section{Appendix}
\label{sec:appendix}

\subsection{Hyperparameters for parameter-efficient fine-tuning}
\label{sec:hyperparam_appendix}
For the prefix tuning experiments, we tune each LLM with the following hyperparameters: 
\begin{itemize}
    \item Initial learning rate of $1e^{-2}$ that linearly decays to $1e^{-3}$ after a 100-step warm-up.
    \item Ten epochs maximum with an early stopping criterion if validation loss has not decayed for five consecutive epochs.
    \item Batch size of eight (large architecture) or 16 (base architecture) with four gradient accumulation steps, rendering an effective batch size of 32 or 64, respectively.
\end{itemize}

For the LoRA experiments, we tune each LLM with the following hyperparameters:
\begin{itemize}
    \item Initial learning rate of $1e^{-3}$ that decays linearly to $1e^{-4}$ after a 100-step warm-up.
    \item Five epochs with no early stopping criterion.
    \item Batch size of six with four gradient accumulation steps, rendering an effective batch size of 24. 
\end{itemize}

LoRA requires slightly more memory than prefix tuning, hence we adjusted the effective batch size to comfortably fit on our NVIDIA Quadro RTX 8000 GPU. Despite the larger memory footprint and greater training time per epoch (Table~\ref{tab:num-params}), LoRA requires fewer epochs to reach convergence than prefix tuning, resulting in 30\% less tuning time overall. We note the importance of a learning rate warm-up over the first 100 gradient steps, which has been shown beneficial for low-data settings~\cite{li2021prefix}. We experimented with various learning rate schedulers (step, exponential decay) but found linear decay to give slightly better performance in terms of validation loss. As discussed in Section~\ref{sec:experiment-details}, the best set of hyperparameters is constant across each of the five models.

\setlength{\tabcolsep}{4pt} 

\begin{table*}
\caption{Quantitative evaluation across each model and adaptation method using the base architecture size. Parameter-efficient (updating <0.4\% of parameters) fine-tuning methods LoRA and prefix tuning drastically outperform discrete prompting strategies. Among these fine-tuning methods, the best performing models are those which have been pretrained on clinical text (\clintfivesci, \clintfive).}
\begin{center}
\begin{tabular}{l l | c c c c}
\textbf{Model} & \textbf{Method} & \textbf{BLEU} & \textbf{ROUGE-L} & \textbf{BERT} & \textbf{F1-Radgraph} \\
\hline
\hline
 & null  & 3.4 & 14.3 & 84.1 & 13.8 \\
 & prefix  & 4.7 & 19.0 & 86.1 & 19.0 \\
 & in-context (1)  & 3.4 & 15.8 & 85.4 & 14.4 \\
\tfive & in-context (2) & 3.3 & 15.8 & 85.4 & 11.8 \\
 & in-context (4)  & 4.4 & 16.2 & 85.5 & 12.1 \\
 & prefix tuning  & 12.9 & 29.1 & 88.4 & 30.7 \\
 & LoRA  & 13.7 & 33.9 & 89.5 & 35.2 \\
\hline
 & null  & 0.5 & 11.3 & 83.0 & 9.7 \\
 & prefix  & 1.1 & 14.7 & 84.7 & 13.8 \\
 & in-context (1)  & 2.9 & 17.8 & 85.6 & 14.6 \\
\flantfive & in-context (2) & 5.3 & 19.6 & 86.2 & 16.6 \\
 & in-context (4)  & 8.6 & 25.0 & 87.0 & 21.6 \\
 & prefix tuning  & 12.1 & 27.1 & 87.8 & 28.0 \\
 & LoRA  & \underline{13.8} & 34.4 & 89.5 & 36.2 \\
\hline
 & null  & 1.0 & 6.4 & 80.0 & 4.2 \\
 & prefix  & 0.3 & 4.2 & 78.0 & 0.7 \\
 & in-context (1)  & 1.8 & 11.3 & 82.0 & 9.7 \\
\scifive & in-context (2) & 2.8 & 12.4 & 82.9 & 12.9 \\
 & in-context (4)  & 3.4 & 12.7 & 83.6 & 14.8 \\
 & prefix tuning  & 10.3 & 28.9 & 88.4 & 30.2 \\
 & LoRA  & 13.5 & 34.6 & 89.6 & 36.1 \\
\hline
 & null  & 1.5 & 7.0 & 78.7 & 6.1 \\
 & prefix  & 1.1 & 5.0 & 77.9 & 4.2 \\
 & in-context (1)  & 0.4 & 9.9 & 73.3 & 7.6 \\
\clintfivesci & in-context (2) & 0.9 & 11.1 & 76.1 & 7.3 \\
 & in-context (4)  & 2.4 & 14.2 & 76.7 & 11.8 \\
 & prefix tuning  & 11.7 & 33.3 & 89.3 & 35.0 \\
 & LoRA  & 13.4 & \underline{36.4} & \underline{89.9} & \underline{37.6} \\
\hline
 & null  & 0.8 & 12.2 & 69.4 & 10.7 \\
 & prefix  & 1.0 & 9.5 & 78.6 & 7.1 \\
 & in-context (1)  & 0.3 & 8.7 & 66.1 & 7.7 \\
\clintfive & in-context (2) & 0.6 & 9.6 & 66.6 & 8.7 \\
 & in-context (4)  & 2.2 & 11.5 & 70.9 & 13.0 \\
 & prefix tuning  & 11.9 & 33.8 & 89.4 & 35.4 \\
 & LoRA  & \textbf{14.8} & \textbf{36.8} & \textbf{89.9} & \textbf{38.2} \\
\end{tabular}
\end{center}
\label{tab:results-base}
\end{table*}
\setlength{\tabcolsep}{4pt} 

\begin{table*}
\caption{Quantitative evaluation on Stanford Hospital's dataset of ultrasound radiology reports with the best adaptation method (LoRA) across each model using the base architecture size. This supports our hypothesis that pretraining with clinical text is beneficial for RRS datasets beyond the MIMIC suite.}
\vspace{-2mm}
\begin{center}
\begin{tabular}{l | c c c c c}
\textbf{Model} & \textbf{BLEU} & \textbf{ROUGE-L} & \textbf{BERT} & \textbf{F1-Radgraph} \\
\hline
\tfive  & 12.6 & 31.2 & 88.2 & 26.2 \\
\flantfive  & 12.0 & 30.6 & 88.3 & 26.8 \\
\scifive  & 13.7 & 30.9 & 88.2 & 26.6 \\
\clintfivesci  & \underline{14.0} & \underline{32.7} & \underline{88.6} & \underline{28.5} \\
\clintfive  & \textbf{15.1} & \textbf{32.8} & \textbf{88.8} & \textbf{29.7} \\
\end{tabular}
\end{center}
\label{tab:results-suh-lora}
\end{table*}

\end{document}